\documentclass[runningheads]{llncs}

\usepackage{eccv}

\usepackage{eccvabbrv}

\usepackage{graphicx}
\usepackage{booktabs}

\usepackage[accsupp]{axessibility}  
\usepackage{makecell}
\usepackage{multirow}
\usepackage{pifont}
\usepackage{fontawesome5}

\usepackage{hyperref}

\usepackage{orcidlink}

\begin{document}

\title{SAM-Med3D: Towards General-purpose Segmentation Models for Volumetric Medical Images} 

\titlerunning{SAM-Med3D}

\author{Haoyu Wang\inst{1,2}\orcidlink{0000-0002-1753-7336} \and
Sizheng Guo\inst{1} \and
Jin Ye\inst{1} \and
Zhongying Deng\inst{1} \and
Junlong Cheng\inst{1} \and
Tianbin Li\inst{1} \and
Jianpin Chen\inst{1} \and
Yanzhou Su\inst{1} \and
Ziyan Huang\inst{1,2} \and
Yiqing Shen\inst{1} \and
Bin Fu\inst{3} \and
Shaoting Zhang\inst{1} \and
Junjun He\inst{1}\orcidlink{0000-0002-1813-1784}$^{*}$ \and
Yu Qiao\inst{1}\orcidlink{0000-0002-1889-2567}\thanks{Corresponding authors.}}

\authorrunning{Wang, H., et al.}

\institute{
Shanghai Jiao Tong University \and
Shanghai Artificial Intelligence Laboratory \and
Shenzhen Institute of Advanced Technology, Chinese Academy of Sciences \\
\email{hejunjun@pjlab.org.cn}
}

\maketitle

\begin{abstract}

Existing volumetric medical image segmentation models are typically task-specific, excelling at specific target but struggling to generalize across anatomical structures or modalities. This limitation restricts their broader clinical use.
In this paper, we introduce SAM-Med3D for general-purpose segmentation on volumetric medical images. Given only a few 3D prompt points, SAM-Med3D can accurately segment diverse anatomical structures and lesions across various modalities.
To achieve this, we gather and process a large-scale 3D medical image dataset, SA-Med3D-140K, from a blend of public sources and licensed private datasets. This dataset includes 22K 3D images and 143K corresponding 3D masks. 
Then SAM-Med3D, a promptable segmentation model characterized by the fully learnable 3D structure, is trained on this dataset using a two-stage procedure and exhibits impressive performance on both seen and unseen segmentation targets. 
We comprehensively evaluate SAM-Med3D on 16 datasets covering diverse medical scenarios, including different anatomical structures, modalities, targets, and zero-shot transferability to new/unseen tasks. The evaluation shows the efficiency and efficacy of SAM-Med3D, as well as its promising application to diverse downstream tasks as a pre-trained model. 
Our approach demonstrates that substantial medical resources can be utilized to develop a general-purpose medical AI for various potential applications.
Our dataset, code, and models are available at: \href{https://github.com/uni-medical/SAM-Med3D}{https://github.com/uni-medical/SAM-Med3D}.
\end{abstract}

\begin{figure*}[tb]
\centering
    \includegraphics[width=\linewidth]{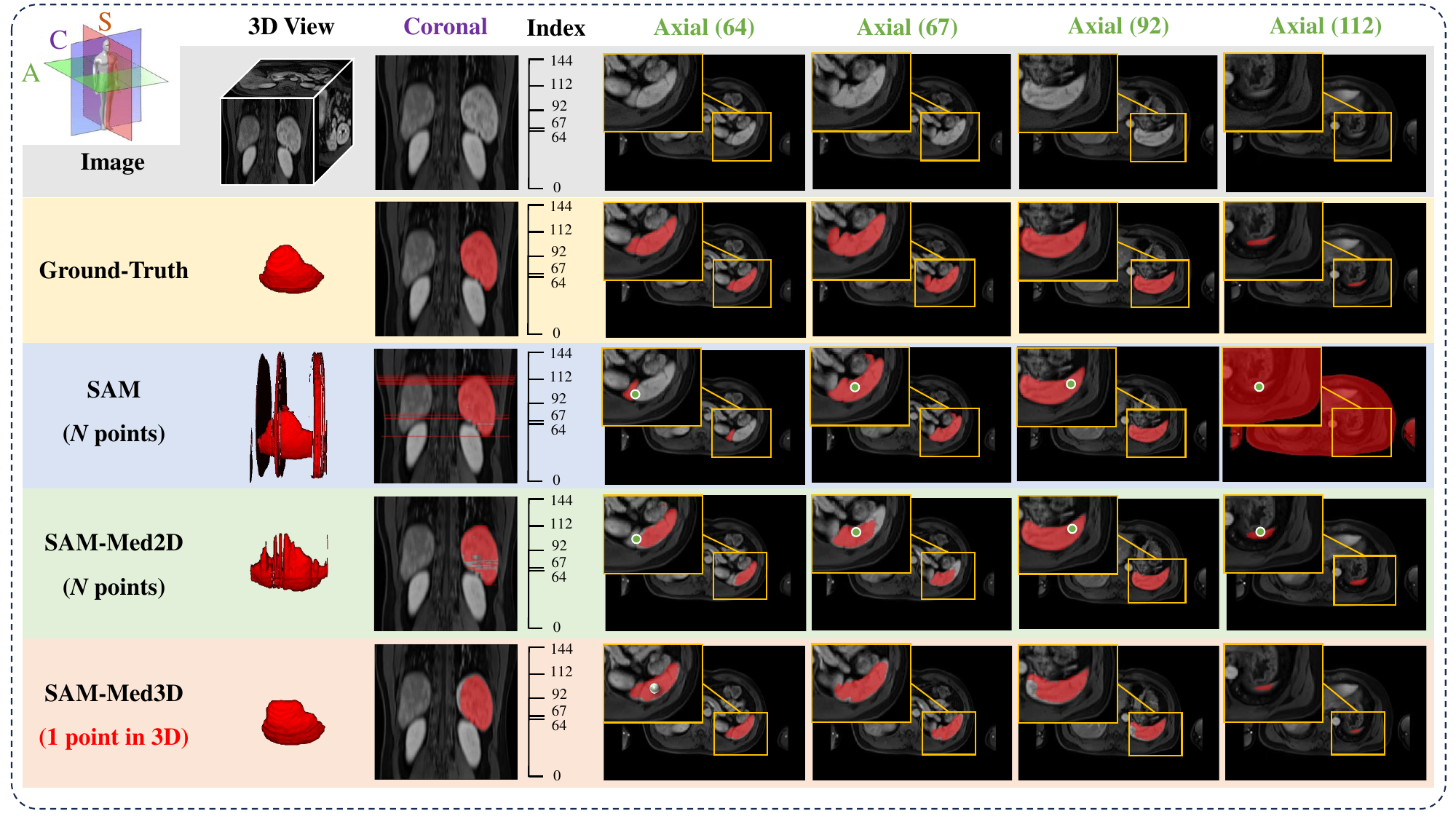}
    \caption{\textbf{Illustration of SAM~\cite{sam}, fine-tuned SAM (SAM-Med2D~\cite{sammed2d}), and our SAM-Med3D on 3D Volumetric Medical Images.} Both SAM and SAM-Med2D take $N$ prompt points (one for each slice) whereas SAM-Med3D uses a single prompt point for the entire 3D volume. Here, $N$ corresponds to the number of slices containing the target object. The top-left corner provides a schematic of the Axial, Coronal, and Sagittal views. For a given 3D input, we visualize the 3D, coronal, and multiple axial views. The numbers in brackets indicate the index of each axial slice.
    }
    \label{fig:motivation}
\vspace{-0.2cm}
\end{figure*}

\section{Introduction}
\label{sec:intro}
Medical image analysis has become an indispensable cornerstone of modern healthcare, aiding disease diagnosis, treatment planning, education, and medical research~\cite{bg1,bg2,stunet}. One of the major challenges in this realm is the precise segmentation of volumetric medical images~\cite{bg6}. Although numerous methods have demonstrated commendable effectiveness across a spectrum of targets~\cite{bg3,bg4,bg5}, mainstream segmentation models tend to specialize in particular organs or lesions. This tendency is attributed to the inherent characteristics of public volumetric medical image datasets: most datasets in medical imaging are collected for specific tasks such as CT images labeled only for organs segmentation~\cite{totalsegmentator}, or PET \& CT images with tumor labels for lesion detection~\cite{gatidis2023autopet}. 
While models trained on such specialized datasets excel in their respective applications, their ability to generalize to other scenarios is limited. Consequently, there arises a necessity to train new models for each specific application.
This approach results in a multitude of task-specific models for different clinical scenarios, which would incur enormous development costs and lead to a significant waste of resources. 

There is a pressing need to build a general-purpose segmentation model for volumetric medical imaging, providing segmentation capabilities applicable to a wide range of anatomical structures and modalities. Currently, Segment Anything Model (SAM)~\cite{sam}, a 2D foundation model built for natural image segmentation, exhibits impressive zero-shot ability to new image distributions and tasks. However, directly applying SAM to 3D volumetric medical images is infeasible, as disclosed in previous evaluations~\cite{mazurowski2023segment,sam3d}. This infeasibility is not only due to the domain gap between medical and natural images but also caused by the 2D structure of SAM failing to capture 3D spatial information in volumetric medical images. 
Here are two straightforward solutions to these issues:

\noindent \textbf{1) Slice-by-Slice Aggregation}~\cite{medsam,sammed2d,medlsam}. This method decomposes each volume into 2D slices, then uses SAM to process each slice, and further aggregates the 2D per-slice results into a 3D prediction. This slice-by-slice manner fails to capture inter-slice correlations, hindering the model's ability to generate consistent 3D predictions. In addition, since each 2D slice necessitates individual prompts for prediction, the large number of 2D slices in a volume yields extensive prompts to obtain the aggregated prediction for such a volume. These prompts, each obtained by per-slice interaction, thereby lead to a burdensome interaction workload.

\noindent \textbf{2) 3D Adapter for Frozen 2D Encoder}~\cite{masam,3dsamAdapt,msa}. In this approach, the 2D encoder layers of SAM are frozen, but 3D adapters are inserted into these frozen 2D layers to enable the model to learn from 3D images. This solution adopts the encoder of SAM to capture 2D information, with adapters only learning about the newly added dimension. This strategy implicitly splits the 3D images along a specific dimension (e.g., two dimensions plus a newly added one), which may curtail their ability to fully model 3D spatial information. 
Besides, existing adapter-based methods~\cite{masam,3dsamAdapt,msa} only use small-scale data with limited target types, which cannot provide the tremendous medical knowledge needed for general-purpose segmentation. Table \ref{tab:method} shows the gap between these methods and our work in dataset size and target diversity.

\begin{table*}[htb]
\caption{\textbf{Comparison of SAM-based models for volumetric medical images.} Our SAM-Med3D employs a fully learnable 3D architecture with large-scale training data, instead of frozen 2D layers with adapters. \textcolor{cyan}{\faSnowflake}~and \textcolor{red}{\faFire*}~denotes frozen and learnable.}
\label{tab:method}
\centering
\resizebox{\textwidth}{!}{
\begin{tabular}{l|ccccc}
\toprule
\textbf{Model} & \makecell{\textbf{Dataset}  \textbf{Size}} & \textbf{Category} & \makecell{\textbf{Image}  \textbf{Encoder}} & \makecell{\textbf{Prompt}  \textbf{Encoder}} & \makecell{\textbf{Mask}  \textbf{Decoder}} \\
\midrule
MedLSAM~\cite{medlsam}                & $\sim$ 25K masks & $\sim$ 50 & \textcolor{cyan}{\faSnowflake}~2D      & \textcolor{cyan}{\faSnowflake}~2D     & \textcolor{cyan}{\faSnowflake}~2D     \\ 

$\mathrm{SAM^{Med}}$~\cite{sammed_nv} & - & - & \textcolor{cyan}{\faSnowflake}~2D      & \textcolor{cyan}{\faSnowflake}~2D     & \textcolor{cyan}{\faSnowflake}~2D     \\ 

SAM3D~\cite{sam3d}                    & 2K masks & 14 & \textcolor{cyan}{\faSnowflake}~2D      & -             & \textcolor{red}{\faFire*}~\textbf{3D}  \\

MA-SAM~\cite{masam}                   & $\leq$ 1K masks & $\leq$ 13 & \textcolor{cyan}{\faSnowflake}~2D + \textcolor{red}{\faFire*}~Adapter   & -             & \textcolor{red}{\faFire*}~2D  \\

MSA~\cite{msa}                        & 12K masks  & 15 & \textcolor{cyan}{\faSnowflake}~2D + \textcolor{red}{\faFire*}~Adapter   &  \textcolor{red}{\faFire*}~2D  & \textcolor{red}{\faFire*}~2D  \\

3DSAM-Adapter~\cite{3dsamAdapt}       & $\leq$ 1K masks & 4 & \textcolor{cyan}{\faSnowflake}~2D + \textcolor{red}{\faFire*}~Adapter   & \textcolor{red}{\faFire*}~\textbf{3D}   & \textcolor{red}{\faFire*}~\textbf{3D}  \\

\midrule
SAM-Med3D                             &  143K masks & 245 & \textcolor{red}{\faFire*}~\textbf{3D}    & \textcolor{red}{\faFire*}~\textbf{3D}   & \textcolor{red}{\faFire*}~\textbf{3D}  \\
\bottomrule
\end{tabular}
}
\end{table*}

\begin{figure}[tb]
\centering
\includegraphics[width=0.7\textwidth]{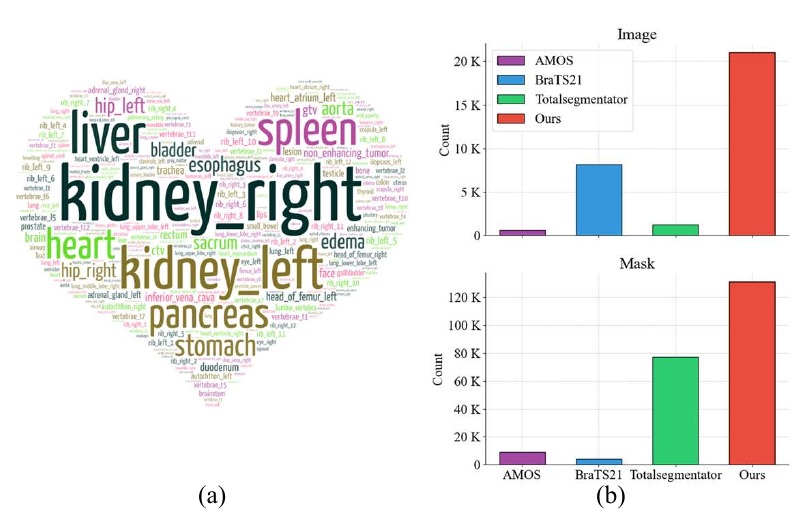}
\caption{\textbf{Overview of SA-Med3D-140K.} (a) The word cloud maps for category statistics of all training data. There are 245 categories in our training data. (b) Comparison of counts of images and masks in the 3D medical image datasets. Our dataset consists of 22K 3D images with corresponding 143K 3D masks, while \textit{AMOS}~\cite{amos}, \textit{TotalSegmentator}~\cite{totalsegmentator} have less than 2K images, and \textit{BraTS21}~\cite{brats2021} has less than 10K masks.} \label{fig:dataset_detail}
\end{figure}

Witnessing the success of SAM, we believe that it is crucial to train a general-purpose 3D medical segmentation model using large-scale and diverse 3D medical data. 
In this paper, we introduce SAM-Med3D, a fully learnable 3D architecture trained on a large-scale volumetric medical dataset using a two-stage procedure. Our SAM-Med3D is a general-purpose promptable segmentation model that needs only a few 3D prompt points to segment diverse anatomical structures and lesions across various modalities, as shown in Fig.~\ref{fig:motivation}. We also highlight that SAM-Med3D is a thorough 3D network with all its modules being 3D so that it can better capture varying 3D spatial patterns in the large-scale volumetric medical dataset. Such dataset, termed SA-Med3D-140K, comprises 22K medical images and 143K masks with 245 categories, derived from an amalgamation of 70 public 3D medical image datasets and 24 private datasets. As evident from Fig.~\ref{fig:dataset_detail}, it is significantly larger than the largest existing public medical image segmentation datasets like \textit{TotalSegmentator}~\cite{totalsegmentator} and \textit{BraTS21}~\cite{brats2021}. In addition, we conducted extensive evaluations to verify the promising performance of SAM-Med3D on both seen and unseen tasks (detailed in Sec.~\ref{sec:quant_eval}). Specifically, we conducted assessments on our SAM-Med3D and its competitors (SAM and SAM-Med2D) utilizing 16 public datasets. We then thoroughly analyze the results from the perspective of anatomical structures, modalities, and categories to evaluate their performance for general-purpose volumetric medical image segmentation. Lastly, we tested the transferability of the SAM-Med3D on multiple downstream segmentation tasks. 

The results reveal three advantages of SAM-Med3D:
\textbf{1) Better interaction efficiency:} SAM-Med3D shows competitive performance in comparison with state-of-the-art methods but requires fewer prompt points (see Fig.~\ref{fig:motivation}). 
This ensures that, compared to previous methods, physicians can utilize SAM-Med3D more readily and enjoy significantly faster speeds than per-slice interaction.  
\textbf{2) Universal segmentation capabilities:} Compared to previous works, SAM-Med3D exhibits broad segmentation capabilities across a wide array of targets and modalities in volumetric medical images. This versatility underscores the strong adaptability and effectiveness of SAM-Med3D in addressing diverse medical imaging challenges.
\textbf{3) Promising transfer ability:} When applied to downstream promptable and semantic segmentation tasks of various new targets, SAM-Med3D exhibits substantial potential as a powerful pre-trained model. 

\section{Related Works}

\textbf{SAM for Volumetric Medical Images.}
Just like Language Foundation Models such as GPT-4, Vision Foundation Models (VFM)~\cite{seggpt,sam,clip,dalle,seem}
have demonstrated powerful generalization capabilities. Among them, SAM~\cite{sam} stands out for promptable image segmentation with zero-shot capabilities in varied visual tasks. Its success in natural images has led to explorations of SAM's potential in volumetric medical imaging, primarily focusing on fine-tuning SAM or incorporating SAM in medical image analysis pipelines. 
MedLSAM~\cite{medlsam} adopts SAM in a two-stage model with a localization model to furnish precise prompt, thereby enhancing SAM's segmentation accuracy. 
SAM3D~\cite{sam3d} leverages a frozen SAM encoder as its image Encoder and  processes voxel images slice-by-slice to procure 3D representations, which are subsequently interpreted through a 3D Decoder to generate masks. 
Besides, other methods try to design 3D adapters to fine-tune SAM for volumetric medical image segmentation. 
3DSAM-Adapter~\cite{3dsamAdapt} and MA-SAM~\cite{masam} have fashioned 3D adapters tailored for each SAM component, transforming the original SAM structure with 3D convolution to facilitate 3D mask formation. In contrast, MSA~\cite{msa} retains all the weights of the original SAM and introduces space and depth adapters specifically designed to process 3D spatial information. Nonetheless, these methods encode the vital 3D information based on adapters~(only partial parameters are trainable) and only fine-tune SAM on specific limited-scale medical datasets.

\noindent \textbf{Evaluation of SAM in Medical Imaging.}
Various studies have explored SAM's effectiveness in medical image segmentation. Cheng et al.~\cite{cheng2023sam} assess SAM across 12 medical datasets using different prompt types and find its performance generally lagging behind state-of-the-art methods. Huang et al.~\cite{huang2023segment} evaluate SAM's zero-shot capability on 52 datasets with three prompt types and observe the subpar performance of SAM. Deng et al.~\cite{deng2023segment} focus on tumor and tissue segmentation, noting SAM's better performance with larger connected entities, while Zhou et al.~\cite{zhou2023sam} identify potential improvements in colonoscopy polyp segmentation using SAM without prompts. Hu et al.~\cite{hu2023sam} investigate SAM in multi-phase liver tumor segmentation in CT scans, finding increased effectiveness with more point prompts. These studies highlight the importance of understanding SAM's capabilities in medical imaging, offering insights to refine SAM for medical segmentation tasks. Complementing these efforts, our research provides a comprehensive evaluation of several SAM-based methods, including our SAM-Med3D, on volumetric medical images.

\section{SA-Med3D-140K Dataset} \label{sec:dataset}
\subsection{Data Acquisition} \label{sec:train_val_set}
We constructed a large-scale 3D volumetric medical image segmentation dataset, SA-Med3D-140K, based on 70 public and 24 private datasets. SA-Med3D-140K contains 22K medical images and 143K masks which is probably the largest volumetric medical image segmentation dataset to date. The dataset covers 28 modalities~(i.e. CT, US, and 26 MR sequences) with 6 major anatomical structures. As shown in Fig.~\ref{fig:dataset_detail}, there are over 240 categories of targets including both organs and lesions in the training set. It is about 20 times larger than the largest well-known medical image segmentation dataset for organ, \textit{TotalSegmentator}~\cite{totalsegmentator}, at the image level. Details of dataset are in Supplementary.

\subsection{Data Pre-processing}
To standardize the diverse data from multiple datasets and control the data quality, we clean and process all the collected data in the following four steps: 

\noindent \textbf{1) Data cleaning based on target shapes:} 
We first summarize the sizes of all targets in each medical image. Then, we removed all masks with a physical size below 1 $cm^3$ or with any single dimension shorter than 1.5 $cm$ to ensure the visibility of target masks under the target spacing 1.5 $mm$.

\noindent \textbf{2) Data cleaning based on volume sizes:} We convert multi-class masks into one-hot formats, calculate the total volume sizes of the foreground, and discard masks where the background exceeds 99\% of the volume.

\noindent \textbf{3) Noise reduction based on connected domains:} For the remaining masks, our pipeline enhances label quality by removing domains smaller than the top-5 largest connected domains.

\noindent \textbf{4) Ambiguity reduction based on symmetry:} We directly manipulate the data by segregating symmetric target masks, such as dividing \textit{Kidney} into \textit{Left Kidney} and \textit{Right Kidney}. This approach contrasts with SAM's method of generating multiple predictions per prompt to address ambiguity; we choose direct data manipulation because we observe that medical image masks typically present less inherent ambiguity.

\subsection{Data Splitting} \label{sec:data_split}
To train and evaluate the performance of our model, we partitioned the SA-Med3D-140K into a training set of 131K masks and a validation set of 12K masks. The training set incorporates 58 public datasets along with all private data. Conversely, the validation set is composed of 16 public benchmark datasets, some of which have their official training sets included within our training set.

This splitting strategy is mainly designed to assess the model's ability to generalize across diverse data source and modalities. Among the 16 datasets in the validation set, all annotated images from 12 datasets have been used for evaluation, implying that 58\% of the cases in our validation set have data sources that are unseen in the training set. Besides, we divided all ultrasound~(US) data into the validation set to assess the cross-modality generalization ability. Details of data splitting are in Supplementary.

\section{Method} \label{sec:method}

\subsection{Devise SAM-Med3D} 
\begin{table}[htb]
\caption{Preliminary experiment comparing three modifications of SAM for volumetric medical images. All models are trained on the training set of \textit{AMOS} and evaluated on seen and unseen targets from the validation/test set of \textit{AMOS} and \textit{Totalsegmentator}. The Dice scores are averaged at prompt points number of 1 and 10 for brevity.}
\label{tab:pre_exp}
    \centering
\begin{tabular}{l |c c | c}
    \toprule
    \multirow{2}{*}{\textbf{Method}}  & \multicolumn{3}{c}{\textbf{Dice (\%)}} \\
    \cline{2-4}
    & Seen & Unseen & Overall\\
\midrule
    3D adapter w/ frozen SAM~\cite{img2video} & 49.08 & 38.60 & 43.84 \\
    Fine-tuning w/ SAM pre-training & 69.13 & 18.43 & 43.78 \\
    Training from scratch on 3D data & 55.79 & 34.63 & 45.21 \\
    \bottomrule
\end{tabular}
    \vspace{-0.4cm}
\end{table}
While SAM excels in prompt-based segmentation on 2D natural images, the 2D architecture of SAM fails to capture 3D spatial information of volumetric images, leading to sub-optimal performance. Modifying SAM for 3D volumetric medical images remains an open question, and there are three potential methods to address this issue:

\noindent \textbf{1) 3D Adapter with Frozen SAM:} Similar to image-to-video adaptation~\cite{img2video}, this approach involves training additional modules~(i.e. adapter) on top of the frozen 2D architecture of SAM to learn 3D medical knowledge.

\noindent \textbf{2) Fine-tuning with SAM Pre-training:} This method repurposes the 2D pre-trained weights for a 3D architecture~(i.e. replicating 2D weights of convolution to formulate 3D weights), and conducts full fine-tuning on 3D medical images to acquire new knowledge.

\noindent \textbf{3) Training from Scratch on 3D Data:} This approach entails designing an entirely 3D structure and training from scratch on 3D volumetric medical data, without utilizing SAM's pre-trained weights.

Each method has its pros and cons. 
The first method benefits from the pre-trained weights of SAM for a speedy convergence, but 3D adapters, built upon the frozen 2D structure of SAM, may be misled by the prior knowledge that is highly biased to 2D natural images. Moreover, most of the additional adapters only learn knowledge for the new z-axis, which is sub-optimal for volumetric medical images with three axes. The second one can also converge fast due to the use of pre-trained weights from SAM. Furthermore, it can directly capture 3D information for its 3D architecture. However, the 2D-to-3D weight transition in this method might further break down the prior knowledge of SAM, which is harmful to the generalization ability the original SAM brings. Such corrupt knowledge may further lead to the lack of strong generalization ability of the fine-tuned model. The third method is advantageous as it is free from biased knowledge from 2D natural images, making it more adaptable for 3D tasks. Yet, it converges much slower and increases the training difficulty.

To identify the optimal solution, we compare them via preliminary experiments. We train a model for each approach on the training set of \textit{AMOS} and evaluate them on both seen and unseen targets, the former from the validation set of \textit{AMOS} and the latter from the test set of \textit{Totalsegmentator}. We average the Dice scores of two prompt settings~(1 and 10 prompt points) for brevity. 

\begin{figure}[tb]
\centering
\includegraphics[width=0.9\textwidth]{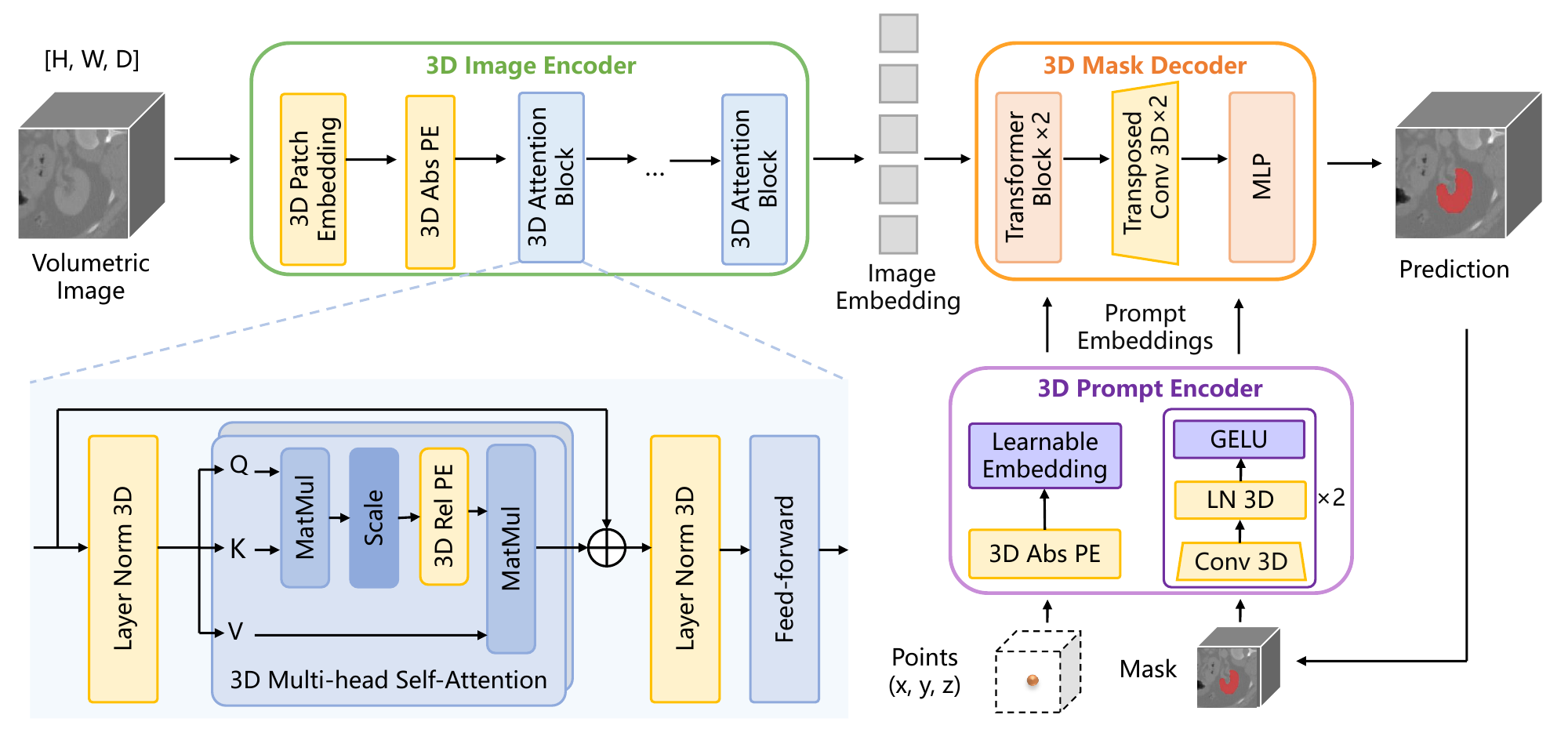}
\vspace{-0.2cm}
\caption{The fully 3D architecture of our SAM-Med3D, encompassing a 3D image encoder, 3D prompt encoder, and 3D mask decoder. 3D positional encoding~(PE) and 3D layers like convolution and layer normalization are employed to construct it.} \label{fig:model_arch}
\vspace{-0.3cm}
\end{figure}

As illustrated in Table \ref{tab:pre_exp}, the 3D adapter shows suboptimal performance on seen targets, while full fine-tuning struggles to generalize on unseen targets; training from scratch emerges as a better trade-off, exhibiting superior average performance. Based on the results, we opted for training a fully 3D architecture from scratch on 3D medical data. The schematics of our SAM-Med3D architecture are in Fig.~\ref{fig:model_arch}. 
Our SAM-Med3D leverages 3D positional encoding and 3D layers (such as convolution and layer normalization) to directly process 3D images and 3D prompts. This pure 3D design can better capture 3D spatial information, and significantly reduce required prompt point numbers. More details of our SAM-Med3D are in Supplementary.

\subsection{Train SAM-Med3D in Two Stages} \label{sec:train_in_two_stage}
Inspired by the prevalence of the pre-training and fine-tuning paradigm in Large Language Models~(LLM), we adopted a two-stage procedure in SAM-Med3D model training. We kept most training setting the same for both stages~(except the iterations), with the major difference being the training data. 

\noindent \textbf{Stage 1: Pre-training.} At the first stage, we utilize all 131K data in the training set of SA-Med3D-140K to construct a powerful pre-trained model. We train the model for 800 epochs until it essentially converges. At this point, SAM-Med3D exhibits performance on common targets that is on par with SAM-Med2D, yet it still falls short on challenging targets.

\noindent \textbf{Stage 2: Fine-tuning.} At this stage, we filter about 75K high-quality masks from the entire training dataset for the 2nd-stage fine-tuning. The impact of different fine-tuning datasets is discussed in Sec.~\ref{sec:transfer_ability}. More details of training are in Supplementary.

\subsection{Evaluate General-Purpose Models} 
Based on the elaborated validation set of the general-purpose dataset SA-Med3D-140K, we undertake a comprehensive evaluation of SAM, SAM-Med2D (the SOTA finetuned SAM for 2D medical images), and our  SAM-Med3D, aiming to set a benchmark for promptable segmentation tasks on 3D medical images. Notably, unlike task-specific methods that train multiple models for various tasks, all the compared methods employ a single model for all different tasks. These models need prompts to yield predictions. In terms of the prompt, we simulated an interactive segmentation clinical scenario using point prompt mode across all models, with the first prompt randomly sampled from the foreground and subsequent points from the error region. 

In our evaluation, SAM-Med2D and SAM resize each slice to target resoltions during the per-slice interaction, while SAM-Med3D uses a patch-based inference for 3D volumes. For volumes over 128, SAM-Med3D starts by cropping a 128-sized patch around the initial point. If the patch's edge has prediction, the model extends its inference in that direction with a 50\% overlapping sliding window, with the prompt still visible at the edge. This approach balances SAM-Med3D's speed and accuracy. Notably, for most targets in our validation set, SAM-Med3D needs to predict only one or two patches.

\section{Experiments} \label{sec:experiment}

\begin{table}[!tb]
\caption{\textbf{Comparison of different methods on our validation set}~(including 16 datasets, detailed in Sec.~\ref{sec:train_val_set}). Here, $N$ denotes the count of slices containing the target~($10 \leq N \leq 200$). $\tau$ represents the interaction time~(usually more than 1 second). }
\label{tab:overall_performance}
    \centering
\resizebox{0.7\textwidth}{!}{
\begin{tabular}{l c |c c c c}
\toprule
\multirow{2}{*}{\textbf{Model}} & \multirow{2}{*}{\textbf{~Prompt~}} & \multirow{2}{*}{\textbf{~Inference Time (s)~}} & \multicolumn{3}{c}{\textbf{Dice (\%)}} \\
\cline{4-6} 
 & & & Seen & Unseen & Overall \\
\midrule

SAM       & $N$  pts & $N(\tau+0.13)$   & 16.79 & 11.73 & 16.15 \\  
SAM-Med2D & $N$  pts & $N(\tau+0.04)$   & 38.91 & 22.55 & 36.83 \\  
SAM-Med3D & 1    pt  & $\tau$+2         & 81.98 & 37.02 & 76.27 \\ 
\midrule
SAM       & 3$N$ pts & 3$N(\tau+0.19)$  & 34.61 & 15.94 & 32.24 \\  
SAM-Med2D & 3$N$ pts & 3$N(\tau+0.07)$  & 51.46 & 29.70 & 48.70 \\  
SAM-Med3D & 3    pts & 3$\tau$+3        & 84.14 & 43.80 & 79.02 \\
\midrule
SAM       & 5$N$ pts & 5$N(\tau+0.25)$  & 49.39 & 21.86 & 45.89 \\  
SAM-Med2D & 5$N$ pts & 5$N(\tau+0.10)$  & 51.89 & 30.41 & 49.17 \\  
SAM-Med3D & 5    pts & 5$\tau$+4        & 84.62 & 46.26 & 79.75 \\  
\midrule
SAM-Med3D & 10 pts     & 10$\tau$+6       & 85.19 & 49.92 & 80.71 \\  

\bottomrule
\end{tabular}
}
    \vspace{-0.3cm}
\end{table}

\subsection{Quantitative Evaluation} \label{sec:quant_eval}
\subsubsection{Overall Performance}

Table \ref{tab:overall_performance} presents the performance of SAM, SAM-Med2D, and SAM-Med3D on our evaluation dataset introduced in Sec.~\ref{sec:data_split}. The first two rows in the table reveal that SAM fine-tuned with medical domain knowledge~(i.e. SAM-Med2D) clearly outperforms SAM by 22.12\% and 10.82\% for seen and unseen targets. 
In contrast, our SAM-Med3D (the 3rd row) shows an even more pronounced performance leap, registering an improvement of 60.12\% in overall Dice scores. Crucially, SAM-Med3D consistently exceeds both SAM and SAM-Med2D across varied prompt point counts, although it needs much fewer points. 

Another significant advantage of our SAM-Med3D is its efficiency. Since the fully 3D architecture of SAM-Med3D enables greater throughput and reduces interaction burden, it operates at only 1\% to 26\% of the inference time required by SAM under different sizes of targets. Even when excluding human interaction time~(i.e. $\tau=0$), SAM costs more inference time when $N$ surpasses 20~($N$ denotes the number of slices containing the target). 
This indicates that for objects with diameters greater than 3 $cm$~(assuming a spacing of 1.5 $mm$), SAM-Med3D emerges as a more efficient choice. This enhancement in efficiency makes it particularly suitable for real-world medical applications where processing larger volumes quickly is crucial.

\begin{table}[!htb]
\caption{\textbf{Comparison of task-specific and general-purpose methods on 6 public benchmarks} (task-specific models are trained on the training set of each dataset). $N$ denotes the count of slices containing the target and $ 10 \leq N $. Items marked with * represent the unseen data source or modality in the training set of SA-Med3D-140K.} 
\label{tab:sota_performance}
\centering
\resizebox{\textwidth}{!}{
\begin{tabular}{l c|c c | c c c c}
\toprule
\multirow{3}{*}{\textbf{Dataset}}  & \multirow{3}{*}{\textbf{Modality}} & \multicolumn{2}{c|}{\textbf{Task-specific}} & \multicolumn{4}{c}{\textbf{General-purpose}} \\
\cline{3-8} & & 
UNETR~\cite{unetr} & nnU-Net~\cite{nnunet} & \makecell{SAM-Med2D~\cite{sammed2d} \\ ($N$ pts)} & \makecell{SegVol~\cite{segvol} \\ (pt+text)} & \makecell{Ours \\ (1 pt)} & \makecell{Ours \\ (10 pts)} \\
\midrule
Totalsegmentator~\cite{totalsegmentator} & CT  &  75.05  &  85.22  & 38.26 & - &  84.68  &  \textbf{87.59} \\ 
KiTS21~\cite{kits2021}           & CT  &  70.75  &  75.32  & 68.74 & - &  72.06  &  \textbf{75.37}  \\ 
AMOS-CT~\cite{amos}          & CT  &  78.33  &  \textbf{88.87}  & 49.61 & - &  79.94  &  83.99  \\ 
AMOS-MR~\cite{amos}          & MR  &  76.29  &  \textbf{86.92}  & 45.53 & - &  75.41  &  81.13  \\ 
BTCV*~\cite{btcv}            & CT  &  78.99  &  81.92  & 50.05 & 73.81 &  79.17  &  \textbf{83.01}  \\ 
TDSC-ABUS23*~\cite{TDSC-ABUS}     & US* &    -    &  45.08  & 49.39 & - &  36.08  &  \textbf{54.35} \\ 
\bottomrule
\end{tabular}
}
    \vspace{-0.3cm}
\end{table}

\subsubsection{Comparison with Task-specific Models}

We conducted a comparison between task-specific models and general-purpose model to show the gap between specialists and generalist on public benchmarks. For task-specific models, we select the representative CNN-based model nnU-Net~\cite{nnunet} and Transformer-based model UNETR~\cite{unetr} to conduct experiments. We trained task-specific models on the training set of each benchmark respectively in our evaluations. For general-purpose models, we tested them using the official weights provided.

Table \ref{tab:sota_performance} summarizes the performance of 5 models on the validation/test set of 6 public benchmarks. We found that SAM-Med3D consistently achieved competitive Dice scores across all datasets compared to task-specific models. When provided with 10 points ($10 \leq N$), SAM-Med3D outperformed all other competitors on 4 benchmarks, including unseen data sources and modalities. We also noted that on the unseen modality Ultrasound~(US), the increase of prompt point numbers can bring more significant performance improvements for SAM-Med3D. The performance of SAM-Med2D also surpasses that of the SOTA 3D task-specific models because its training data includes 2D US slices.

\subsubsection{Evaluation on Different Anatomical Structures} \label{sec:eval_anat}

\begin{table*}[htb]
\caption{Comparison from the perspective of anatomical structure and lesion on our validation set~(including 16 datasets, detailed in Sec.~\ref{sec:train_val_set}). Abd\&Tho represents Abdominal and Thorax targets. $N$ denotes the count of slices containing the target~($10 \leq N \leq 200$).}
\label{tab:anatomical_performance}
\centering
\resizebox{0.9\textwidth}{!}{
\begin{tabular}{l c |c c c c c c | c c}
\toprule
\multirow{2}{*}{\textbf{Model}} & \multirow{2}{*}{\textbf{Prompt}} & \multicolumn{6}{c|}{\textbf{Seen}} & \multicolumn{2}{c}{\textbf{Unseen}} \\
\cline{3-10}
& & Abd\&Tho & Bone & Brain & Cardiac & Muscle & Lesion & Organ & Lesion  \\
\midrule
SAM & $N$         pts & 19.93 & 17.85 & 29.73 & 8.44  & 3.93  & 11.56 & 12.14 & 8.88  \\
SAM-Med2D & $N$   pts & 50.47 & 32.70 & 36.00 & 40.18 & 43.85 & 24.90 & 19.36 & 44.87 \\
SAM-Med3D & 1     pt  & 80.76 & 83.38 & 43.74 & 87.12 & 89.74 & 58.06 & 35.99 & 44.22 \\
\midrule
SAM & 3$N$        pts & 31.97 & 42.07 & 40.42 & 18.94 & 8.46  & 24.52 & 16.18 & 14.21 \\
SAM-Med2D & 3$N$  pts & 58.94 & 47.48 & 54.57 & 47.06 & 56.05 & 42.60 & 27.25 & 46.87 \\
SAM-Med3D & 3     pts & 84.06 & 84.74 & 53.34 & 89.71 & 91.01 & 61.73 & 42.20 & 55.02 \\
\midrule
SAM & 5$N$        pts & 43.45 & 59.87 & 49.85 & 33.57 & 16.94 & 38.86 & 21.67 & 23.22 \\
SAM-Med2D & 5$N$  pts & 59.52 & 47.81 & 55.02 & 47.38 & 56.36 & 43.90 & 27.95 & 47.66 \\
SAM-Med3D & 5     pts & 84.68 & 85.02 & 56.86 & 90.36 & 91.29 & 62.94 & 44.51 & 58.46 \\
\midrule
SAM-Med3D & 10 pts   & 85.42 & 85.34 & 61.27 & 90.97 & 91.62 & 64.80 & 48.10 & 62.72 \\
\bottomrule
\end{tabular}
}
    \vspace{-0.3cm}

\end{table*}

\begin{figure*}[htb]
    \centering
    \includegraphics[width=\textwidth]{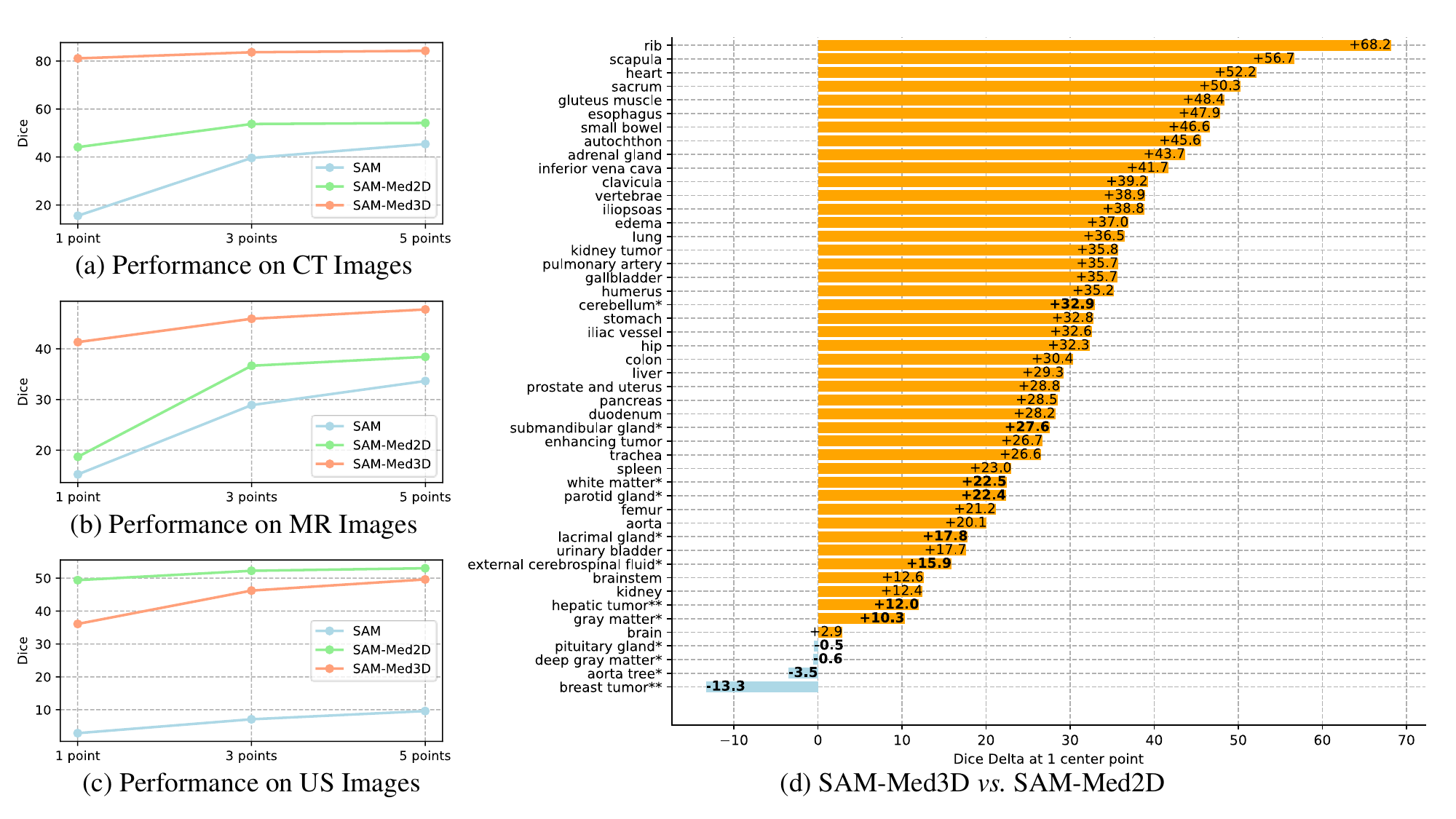}
    \vspace{-.6cm}
    \caption{(a-c) Comparison across different modalities with varying numbers of points. Despite not being trained on the US modality like SAM-Med2D, SAM-Med3D still shows competitive performance. (d) Comparison of the Dice score between SAM-Med3D and the 2D fine-tuned SAM, SAM-Med2D~\cite{sammed2d} across 44 major organs and 5 kinds of lesions. $*$ and $**$ represent unseen organs and lesions.}
    \label{fig:sota_compare}
\end{figure*}

Table \ref{tab:anatomical_performance} summarizes the results of SAM, SAM-Med2D, and our SAM-Med3D across different anatomical structures and lesions. With only $N$ points (1 point/slice), SAM frequently exhibits poor performance. This issue is not readily resolved even with additional points ($3N$ or $5N$ points), especially in anatomical structures with unclear boundaries such as cardiac and muscle regions. In contrast, SAM-Med2D and our SAM-Med3D, infused with medical knowledge, are more effective in accurately identifying targets with just a single point per slice or volume. Notably, with only one point, SAM-Med3D achieves significant better performance than SAM-Med2D across a range of seen and unseen targets, except the unseen lesion. This outlier occurs because one of the unseen lesion modality~(i.e. Ultrasound) for SAM-Med3D, is already familiar to SAM-Med2D. As the number of prompt points is increased, our SAM-Med3D distinctly leads for all examined anatomical structures and lesions, showcasing its robustness and versatility. 

In our evaluation of seen targets across various anatomical structures, we found that segmenting the brain structures and lesions was more challenging for all models. We noted that for the brain structures, with sufficient prompts, SAM-Med2D begins to perform competitively with SAM-Med3D; however, for lesions, additional prompts did not significantly assist SAM-Med2D in reducing the gap with SAM-Med3D. We attribute this difference to lesions being more dependent on prompts. This idea is corroborated in the case of unseen targets. In comparison to unseen organs, the increase in prompt points has a significantly larger effect on SAM-Med3D's recognition of unseen lesions.

\subsubsection{Evaluation on Different Modalities}

We compared the three methods for three dominant modalities in volumetric medical imaging~(CT, MR, and US) across various numbers of prompt points. 
For brevity, we averaged the results from all MR sequences to present them effectively as MR results. 
As shown in Fig.~\ref{fig:sota_compare}, SAM underperforms in segmenting all three volumetric modalities with just one point per slice, yielding Dice scores below 20\%. Although SAM shows better results on all three modalities with more prompt points for each slice, its performance is still far away from the other two methods. 

It is observed that SAM-Med3D performs better with two mainstream imaging modalities~(CT and MR). Even in the case of the unseen modality US, SAM-Med3D exhibits comparable performance when provided with sufficient prompts.

\subsubsection{Evaluation on Major Organs and Lesions}
We have organized all the categories and classified them according to the specific organ or lesion that they belong to. For example, the results for \textit{Left Kidney} and \textit{Right Kidney} are averaged into \textit{Kidney}. By this means, we have identified 44 major organs and 5 kinds of lesions.
Fig.~\ref{fig:sota_compare} (d) shows that SAM-Med3D using 1 point outperforms SAM-Med2D with $N$ points in 45 targets out of 49, achieving up to +68.2\% improvement. 
Consistent with the observation of anatomical structures in Sec.~\ref{sec:eval_anat}, SAM-Med3D shows greater advantages in segmenting bones, cardiac structures and muscles, maybe benefiting from their sensitivity to the 3D spatial correlations. For brain structures and unseen targets, SAM-Med3D exhibits a relatively smaller advantage.

\begin{table}[htb]
\caption{\textbf{Transferability evaluation of semantic segmentation}. We trained the SOTA ViT-based model~(i.e. UNETR~\cite{unetr}), with and without our SAM-Med3D pre-trained ViT encoder, to assess the benefits of pre-training.}
\vspace{-0.3cm}
\label{tab:transfer_performance}
    \centering
\resizebox{0.9\textwidth}{!}{
    \begin{tabular}{c|cccc|c}
        \toprule
        \textbf{Pre-train} & \textbf{AMOS~\cite{amos}} & \textbf{Totalsegmentator~\cite{totalsegmentator}} & \textbf{CAS2023~\cite{CAS}} & \textbf{SEG.A.2023~\cite{SEG.A.}} & \textbf{Avg.}\\
        \midrule
        -          & 76.29 & 82.67 & 86.34 & 87.05 & 83.09 \\
        SAM-Med3D  & 81.92 & 85.17 & 88.39 & 87.57 & 85.79 \\
        \bottomrule
    \end{tabular}
}
\vspace{-0.5cm}
\end{table}

\begin{table*}[htb]
\centering
\caption{\textbf{Transferability evaluation of promptable segmentation}. The first row displays baseline results from SAM-Med3D only with the 1st-stage pre-training. The impact of various fine-tuning datasets during the 2nd-stage fine-tuning is assessed on our validation set~(excluding \textit{FeTA22}~\cite{feta} for fairness). The last row denotes SAM-Med3D fine-tuned with high-quality data selected from 44 datasets in SA-Med3D-140K.}
\vspace{-0.3cm}
\label{tab:anatomical_performance_ft}
\resizebox{\textwidth}{!}{
\begin{tabular}{c c | c c c c c c c | c c}
\toprule
\multirow{3}{*}{\makecell{\textbf{Fine-tune}\\ \textbf{Target}}}& \multirow{3}{*}{\makecell{\textbf{Fine-tune}\\ \textbf{Dataset}}} & \multicolumn{7}{c|}{\textbf{Seen}} & \multicolumn{2}{c}{\textbf{Unseen}} \\
\cline{3-11}
& & Abd\&Tho & \makecell{Bone\\(vertebrae)} & \makecell{Bone\\(other)} & Brain & Cardiac & Muscle & Lesion & Organ & Lesion  \\
        \midrule

-       & -                               & 58.61 & 66.05 & 28.08 & 56.34 & 68.50 & 69.45 & 47.87 & 27.80 & 48.44 \\
organ & \textit{AMOS}~\cite{amos}       & 67.23 & 58.59 & 30.51 & 59.72 & 71.89 & 77.39 & 50.47 & 38.30 & 37.43 \\
vertebrae & \textit{Verse20}~\cite{verse} & 35.08 & 78.84 & 23.77 & 7.77  & 61.02 & 37.75 & 35.42 & 18.38 & 24.69 \\
brain & \textit{FeTA22}~\cite{feta}       & 50.28 & 51.14 & 25.34 & 62.56 & 62.95 & 43.70 & 45.89 & 44.81 & 24.73 \\
all & 44 datasets                         & 85.42 & 85.98 & 84.90 & 61.91 & 90.97 & 91.62 & 64.80 & 48.10 & 62.72 \\

\bottomrule
\end{tabular}
}
\vspace{-0.3cm}
\end{table*}

\subsubsection{Evaluation on Transferability} \label{sec:transfer_ability}


We test the transferability of SAM-Med3D as a pre-train model on two important downstream tasks:

For semantic segmentation, we choose two frequently-used benchmarks~(i.e. \textit{AMOS}~\cite{amos}, \textit{Totalsegmentator}~\cite{totalsegmentator}) and two unseen datasets from the MICCAI 2023 Challenge~(i.e. \textit{CAS2023}~\cite{CAS}, \textit{SEG.A.2023}~\cite{SEG.A.}). The image encoder of SAM-Med3D is used as the pre-trained feature extractor for UNETR~\cite{unetr}, a leading-edge ViT-based model in medical segmentation tasks. As depicted in Table~\ref{tab:transfer_performance}, when fine-tuned with pre-trained ViT encoder from SAM-Med3D, UNETR demonstrates a substantial performance boost compared to these without pre-training, achieving a maximum improvement of 5.63\% in the Dice score. Consequently, SAM-Med3D demonstrates the substantial potential to establish itself as a powerful pre-trained ViT encoder for multiple downstream tasks. 


For promptable segmentation, we carried out experiments as outlined in Section \ref{sec:train_in_two_stage}. We assessed the impact of various fine-tuning datasets during the second-stage fine-tuning from the viewpoint of anatomical structures and lesions. As demonstrated in Table~\ref{tab:anatomical_performance_ft}, fine-tuning on specific datasets significantly enhances the model's expertise in targeted entities; however, this specialization inadvertently biases the model, leading to decreased performance on non-target entities. Conversely, fine-tuning on carefully selected, high-quality data further strengthens the model's general-purpose segmentation performance. 


\subsection{Qualitative Evaluation} \label{sec:vis}

\begin{figure*}[tb]
\centering
\includegraphics[width=0.9\linewidth]{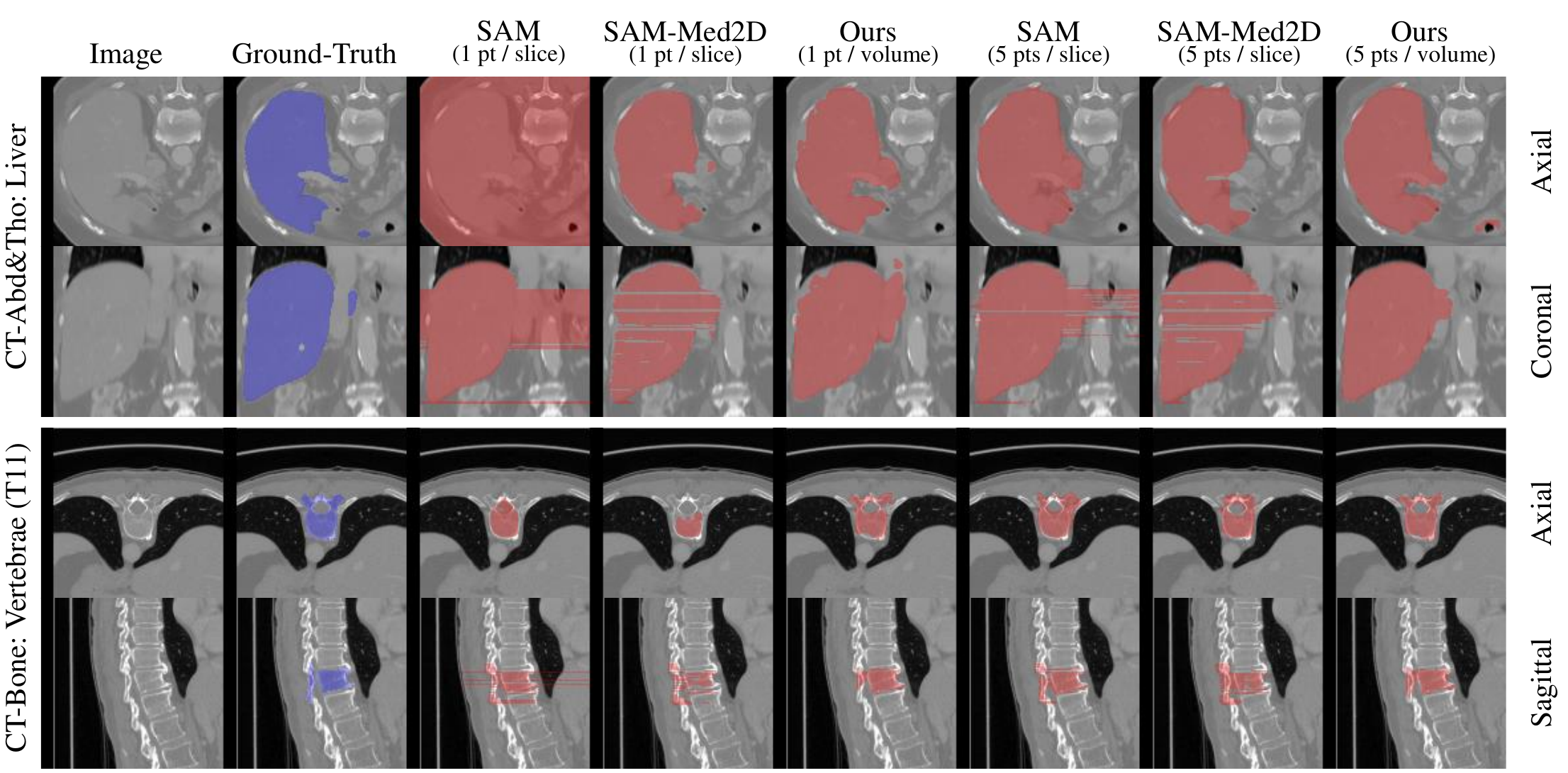}
\vspace{-0.3cm}
\caption{Visualization of SAM, SAM-Med2D, and our SAM-Med3D across diverse anatomical structures and modalities for 1 or 5 points. We present both axial and coronal/sagittal views to illustrate the 3D results comprehensively. } 
\label{fig:vis_all}
\vspace{-0.3cm}
\end{figure*}

To qualitatively compare the performance of SAM-Med3D and other methods, we visualize the predicted masks across different point numbers in Fig.~\ref{fig:motivation} and \ref{fig:vis_all}. 
Based on these visualization results, we highlight two key observations: 1) SAM-Med3D requires fewer prompts. 2) SAM-Med3D shows better inter-slice consistency than all other methods.


\section{Conclusion}
In this paper, we introduce a general-purpose promptable model for volumetric medical image segmentation, dubbed SAM-Med3D. SAM-Med3D uses only a few prompt points to correctly segment both seen and unseen anatomical structures and modalities, underpinning its general-purpose segmentation ability. Such an impressive ability is obtained by training this fully learnable 3D network on a large-scale 3D medical dataset in two stages. Extensive evaluations on 16 public datasets verify the efficacy and efficiency of our SAM-Med3D in various clinical settings, e.g., a 60.12\% improvement over SAM with one point per volume. Our model also shows promising potential to serve as a powerful pre-trained transformer model for multiple downstream segmentation tasks.

%
%
\bibliographystyle{splncs04}
\bibliography{main}
\end{document}